\documentclass[letterpaper, 10 pt, conference]{ieeeconf}  
   \makeatletter
   \let\NAT@parse\undefined
   \makeatother

\IEEEoverridecommandlockouts                              

\usepackage{url}
\usepackage{amsmath,amssymb,amsfonts}%
\usepackage{subcaption}
\usepackage{graphicx}  
\usepackage{framed} 
\usepackage{graphicx}%
\usepackage{multirow}%
\usepackage{mathrsfs}%

\usepackage{xcolor}%
\usepackage{textcomp}%
\usepackage{manyfoot}%
\usepackage{booktabs}%
\usepackage{algorithm}%
\usepackage{algorithmicx}%
\usepackage{algpseudocode}%
\usepackage{listings}%
\usepackage{lmodern}
\usepackage{lipsum}
\usepackage[numbers]{natbib}
 
\title{\LARGE \bf
Collecting Larg-Scale Robotic Datasets\\ on a High-Speed Mobile Platform
}

\makeatletter
\newcommand{\linebreakand}{%
  \end{@IEEEauthorhalign}
  \hfill\mbox{}\par
  \mbox{}\hfill\begin{@IEEEauthorhalign}
}
\makeatother

\author{Yuxin Lin \and Jiaxuan Ma \and Sizhe Gu \and Jipeng Kong \and Bowen Xu \linebreakand Xiting Zhao \and Dengji Zhao \and Wenhan Cao \and S\"oren Schwertfeger$^{1}$
	\thanks{$^{1}$ All authors are with the Key Laboratory of Intelligent Perception and Human-Machine Collaboration -- ShanghaiTech University, Ministry of Education, China;
        {\tt\small \{linyx2023, majx2023, guszh2023, v-kongjp, xubw, zhaoxt, zhaodj, whcao, soerensch\}@shanghaitech.edu.cn}}%
}

\begin{document}

\maketitle
\thispagestyle{empty}
\pagestyle{empty}

\begin{abstract}
Mobile robotics datasets are essential for research on robotics, for example for research on Simultaneous Localization and Mapping (SLAM). Therefore the ShanghaiTech Mapping Robot was constructed, that features a multitude high-performance sensors and a 16-node cluster to collect all this data. That robot is based on a Clearpath Husky mobile base with a maximum speed of 1 meter per second. This is fine for indoor datasets, but to collect large-scale outdoor datasets a faster platform is needed. This system paper introduces our high-speed mobile platform for data collection. The mapping robot is secured on the rear-steered flatbed car with maximum field of view. Additionally two encoders collect odometry data from two of the car wheels and an external sensor plate houses a downlooking RGB and event camera. With this setup a dataset of more than 10km in the underground parking garage and the outside of our campus was collected and is published with this paper. \end{abstract}

\section{Introduction}

The ShanghaiTech Mapping Robot \cite{xu2024shanghaitech} is a powerful system to collect sensor data. It features many sensors, such as 5 stereo pairs of GS3-U3-51S5C-C 5MP RGB cameras with up to 60Hz frame rate, a Ladybug5+ omnidirectional camera system, 5 RGB-D and 5 radar sensors, 7 LiDARs, stereo infrared cameras, stereo event cameras, IMUs and an RTK GPS system. All this data is collected with a 16-node "Cluster on Wheels" \cite{yang2022cluster}. The goal of this mapping robot is to collect universal datasets for research on robotic algorithms such as Simultaneous Localization and Mapping (SLAM) \cite{cadena2016past}. For that we will use SLAM Hive \cite{liu2024benchmarking, yang2023slam}, a cloud-based system for benchmarking and analysing 10s of thousands of mapping runs. 

The mobile base of the ShanghaiTech Mapping Robot is a Clearpah Husky. This robot has a maximum speed of 1m/s. This makes this system impractical for collecting large scale datasets, e.g. of the big parking garage or the campus of our school. Recognizing this challenge, this paper proposes a novel approach to enhance the mapping robot's speed and efficiency by leveraging the mobility of a small vehicle. By securing the mapping robot to a flatbed car, we aim to significantly improve its ability to cover large distances quickly and efficiently. This solution not only addresses the issue of speed but also allows for the strategic placement of sensors, such as downward-looking cameras, on this vehicle to optimize data collection quality, while maintaining the important sensor field of view in the forward direction.

The contributions of this paper are:
\begin{itemize}
    \item The design and implementation of a secure mounting mechanism for the wheeled mapping robot on the flatbed car.
    \item The design and implementation of an external dual-wheel encoding system to record odometry data.
    \item The design of an external downlooking sensor platform connected to the ShanghaiTech Mapping Robot.
    \item The collection and sharing of a large-scale dataset of the ShanghaiTech underground and campus area, together with all code and CAD data.
\end{itemize}

\begin{figure}[tb]
    \centering  
    \includegraphics[width=1\linewidth]{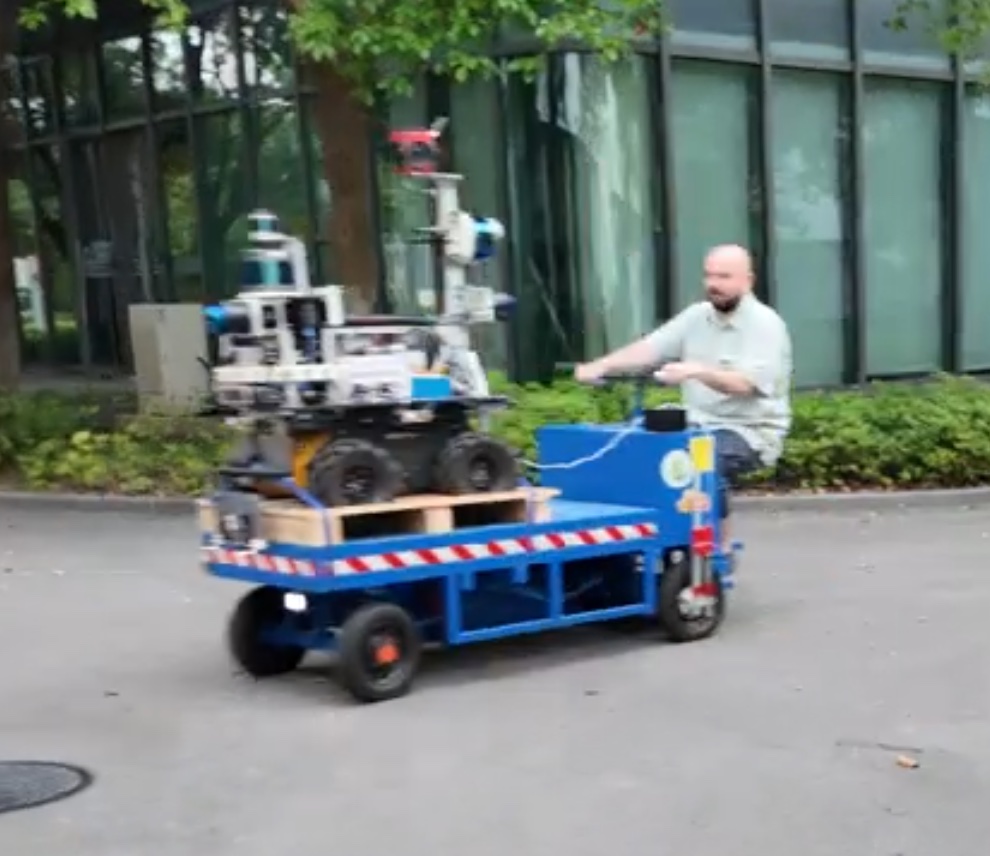}
        \caption{Flatbed car with ShanghaiTech Mapping Robot, encoders and extra downlooking cameras.}
    \label{fig:experiment}
    \vspace{-5mm}
\end{figure}

The rest of the paper is organized as follows: Section \ref{sec:state} introduces the state of the art of dataset collection before the ShanghaiTech Mapping Robot. The hardware and software we designed for the flatbed mapping robot system is described in Section \ref{sec:system}. That includes the mounting system, wheel encoder setup as well as an external downlooking sensor plate. The dataset and the SLAM experiment with it are presented in Section \ref{sec:dataset}. The paper concludes with Section \ref{sec:conclusions}.
\section{State of the Art}
\label{sec:state}
Datasets, such as the one to be collected with our platform, play an important role in robotics. The Oxford Radar RobotCar Dataset~\cite{9196884} is a pioneering dataset designed for researching scene understanding using Millimetre-Wave FMCW scanning radar data. This dataset targets autonomous vehicle applications, leveraging the robustness of radar technology against adverse environmental conditions such as fog, rain, snow, or lens flare, which often pose challenges to other sensor modalities like vision and LiDAR. Collected in January 2019, the dataset encompasses thirty-two traversals of a central route in Oxford, covering a total distance of 280 km of urban driving.

One of the most prominent car datasets is the KITTI Dataset \cite{geiger2013vision}. It features RGB and LiDAR data on public roads. In contrast, we aim to also collect data in places inaccessible to fully sized cars. Instead we are more interested in the mobile robotics domain. 

In the realm of high-precision mapping, the Advanced Mapping Robot and High-Resolution Dataset~\cite{CHEN2020103559} introduces a fully hardware-synchronized mapping robot, which includes support for an externally synchronized tracking system to achieve super-precise timing and localization. The mapping robot is equipped with nine high-resolution cameras and two 32-beam 3D LiDARs, complemented by a professional static 3D scanner for ground truth map collection. With comprehensive sensor calibration, three distinct datasets were acquired to evaluate the performance of mapping algorithms within and between rooms.

The S3E dataset~\cite{feng2023s3e} offers a novel large-scale multimodal collection captured by a fleet of unmanned ground vehicles following four collaborative trajectory paradigms. This dataset comprises 7 outdoor and 5 indoor scenes, each exceeding 200 seconds in duration, featuring well-synchronized and calibrated high-quality stereo camera, LiDAR, and high-frequency IMU data. S3E surpasses previous datasets in terms of size, scene variability, and complexity, boasting four times the average recording time compared to the pioneering EuRoC dataset.

LiDAR-based SLAM for Robotic Mapping: State of the Art and New Frontiers~\cite{Yue2024LiDARbasedSF} provides an exhaustive literature survey on LiDAR-based simultaneous localization and mapping (SLAM) techniques. This comprehensive review covers various LiDAR types and configurations, categorizing studies into 2D LiDAR, 3D LiDAR, and spinning-actuated LiDAR. The paper highlights the strengths and weaknesses of these systems and explores emerging trends such as multi-robot collaborative mapping and the integration of deep learning with 3D LiDAR data to enhance SLAM performance in complex environments.

Reconstructing maps using data obtained from six degrees of freedom poses significant challenges due to the asynchronous nature of ranging data reception, leading to potential mismatches in the generated point clouds~\cite{zhang2017low}. The SegMatch method addresses this challenge by matching 3D segments, thereby enhancing localization robustness without relying on perfect segmentation or predefined objects~\cite{7989618}.

LeGO-LOAM~\cite{zhang2017low} is a lightweight, real-time six-degree-of-freedom pose estimation method specifically designed for ground vehicles operating on varying terrain. This method reduces noise through point cloud segmentation and feature extraction, utilizing planar and edge features for a two-step Levenberg-Marquardt optimization to solve transformation problems between consecutive scans, thereby improving both accuracy and computational efficiency.

To the best of our knowledge, we are the first researchers attempting to mount a complete mapping robot on an additional faster platform in order to enhance its operation speed and range.

\section{System Description}
\label{sec:system}

An electric flatbed car has been selected as our high-speed mobile platform to accommodate our mapping robot, thus overcoming the speed limitations of the current mapping robot. Specifically, we have chosen the hydraulic lifting flatbed car model. This vehicle meets our requirements for load platform speed and endurance, with an empty car range of 30 km and a maximum speed of 25 km/h. The motor has a power rating of 1000W, which is sufficient for our needs. It mounts the robot at the front, while the driver is, with a very small footprint, sitting at the back, thus giving full field-of-view to the front and sides and only obstructing a little to the back (depending on the drivers size :/).

The lifting platform of this vehicle is beneficial for our load handling tasks, as well as for the information collection and map construction tasks performed by the robot. The size of the load platform is 1500 x 800 mm, and it has a lifting capacity of 500 kg. Additionally, the platform can be raised to a maximum height of 1600 mm above the ground.

\begin{figure}[htbp]
    \centering  
        \includegraphics[width=0.5\textwidth]{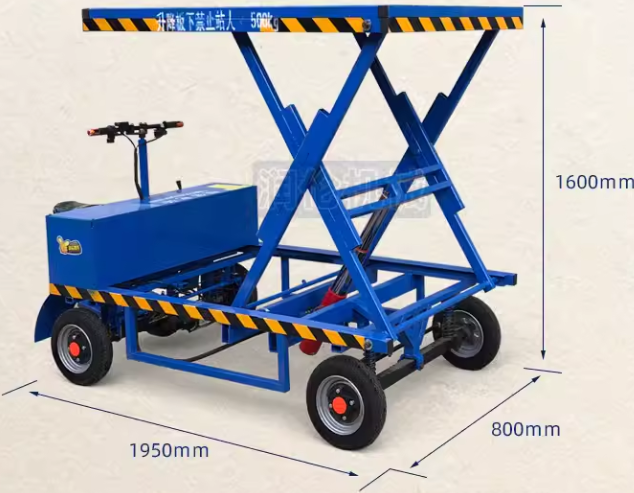}
        \caption{The Hydraulic lifting flatbed car.}
\end{figure}

\begin{figure}[t]
    \centering  
        \includegraphics[width=0.5\textwidth]{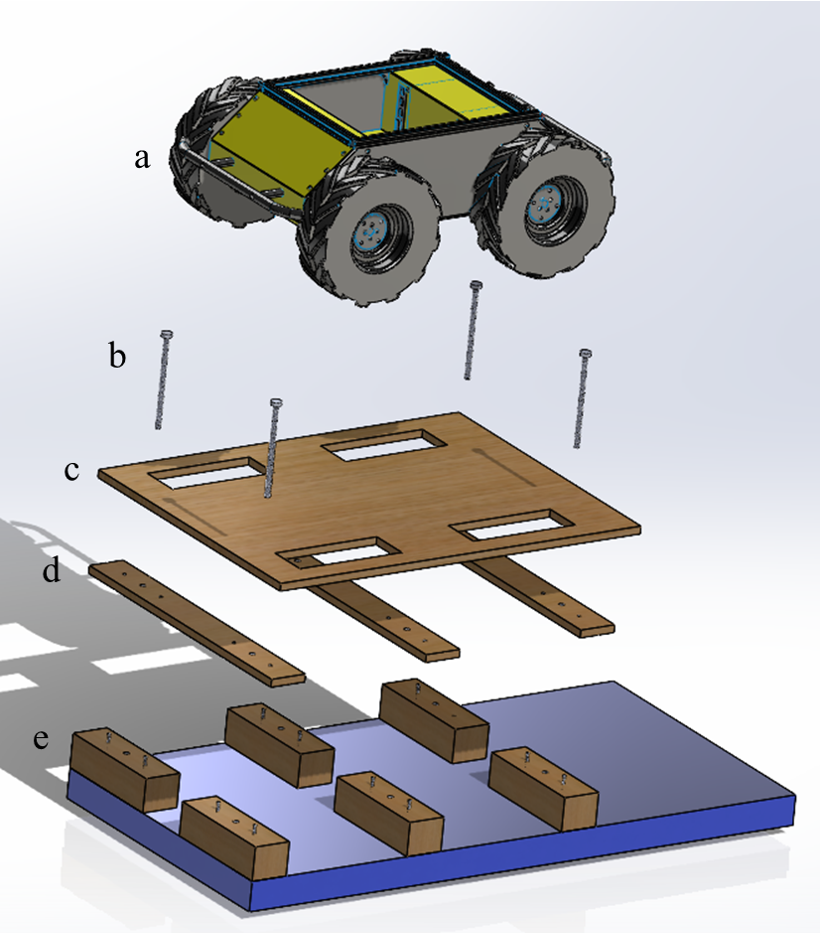}
        \caption{Explode vision.  a) UGV. b) Long bolts. c) wood plate. d) wooden strips. e) wooden beam.}
    \label{fig:explode_vision}
\end{figure}

\begin{figure}[b!]
    \centering  
        \includegraphics[width=0.5\textwidth]{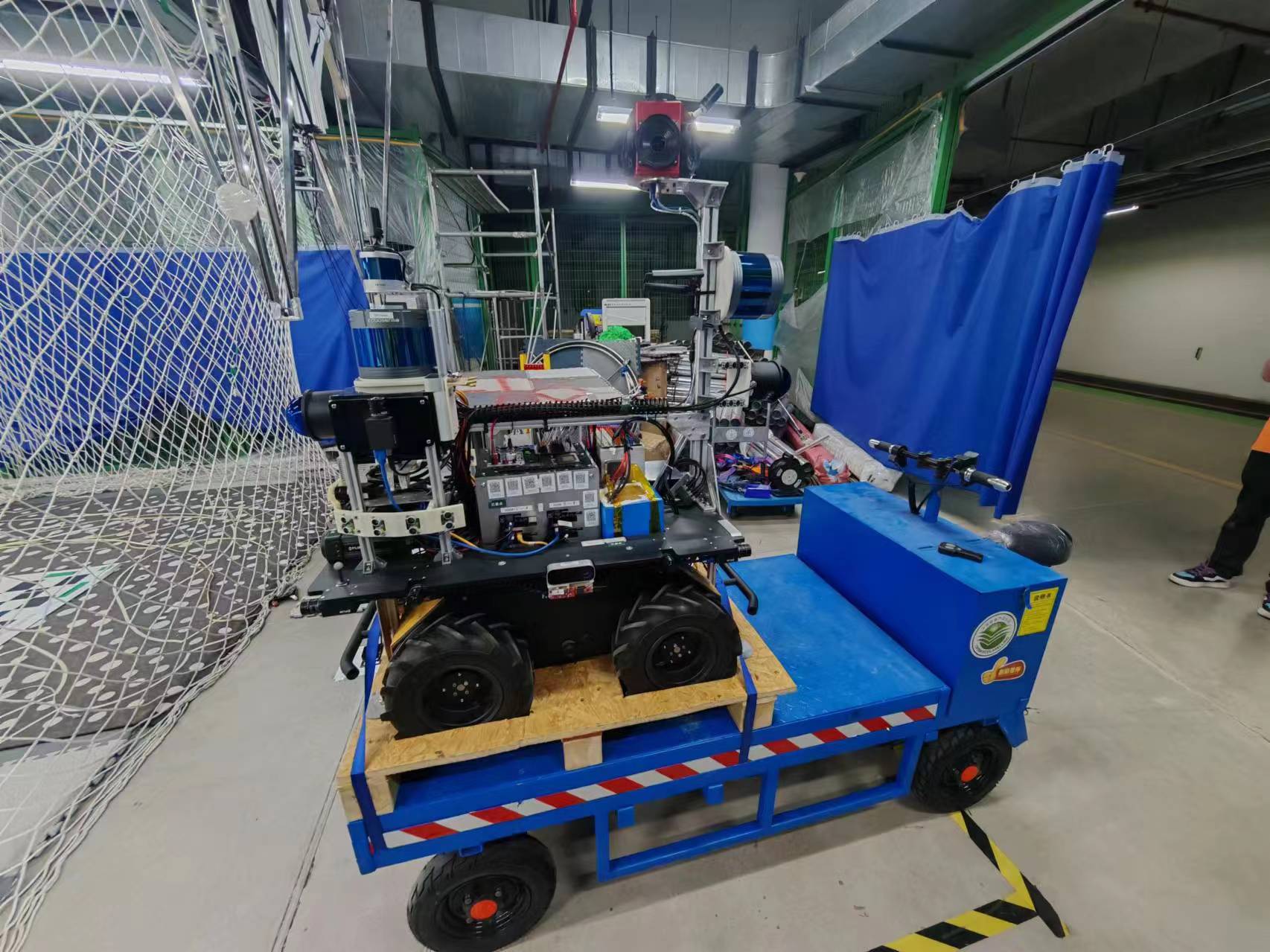}
        \caption{Car with frame prototype.}
    \label{fig:car_with_frame}
\end{figure}

\subsection{Mounting System}

The next challenge we face is how to securely mount the mapping robot onto the hydraulic lifting flatbed car. Our proposed solution involves using a wood plate with a thickness of 21mm as the top layer of the structure. This plate will have four holes to accommodate the robot's wheels. To ensure stability, six wooden beams will be used to connect the wood plate to the flatbed structure of the car, elevating the entire assembly. Each pair of wooden beams will be reinforced with wooden strips to provide additional support. Finally, the entire structure will be firmly secured using long bolts, which will be fastened from top to bottom. This method ensures that the mapping robot is securely and stably mounted onto the hydraulic lifting flatbed car. (Fig.~\ref{fig:explode_vision})

Based on the actual conditions, we adjusted the parameters and shape of the frame. As length of the hole is less than the diameter of the wheel, the elasticity of the rubber tire and the texture on the wheel's surface allow the tire to fit securely into the hole, thereby preventing the safety issues associated with tire rolling during travel. The placement process requires ensuring the vehicle is level to avoid reading errors caused by tilting. The prototype is shown in Fig.~\ref{fig:car_with_frame}.


\subsection{Wheel Odometry}

\begin{figure}[htbp]
    \centering  
    \includegraphics[width=0.5\textwidth]{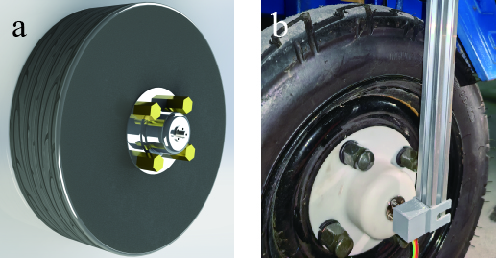}
    \caption{The scheme for installing a bracket on the wheel. a) Adding a bracket at the center of the wheel. b) Locking the rotation of the wheel and the encoder shaft through the bracket and a flange coupling.}
    \label{fig:existing_plan}
\end{figure}

During actual driving, it is necessary to obtain the rotation angles of the vehicle's left and right wheels to calculate the travel distance and direction. We achieve this by connecting an encoder to each of the two passive back wheels and locking their rotation together, allowing the encoders to capture the relevant wheel rotation information. Due to the vehicle's structure not meeting the encoder installation requirements for this task, we need to design an additional framework to mount the encoder while ensuring it rotates co-axially with the wheel. Our approach involves installing a platform at the wheel's center, secured by four bolts originally surrounding the tire. This ensures coaxial rotation of the platform and the wheel. A flange coupling can be installed on top of the platform to lock the rotation of the encoder shaft and the platform. The setup is illustrated in Fig.~\ref{fig:existing_plan}.

\begin{figure}[htbp]
    \centering  
    \includegraphics[width=0.5\textwidth]{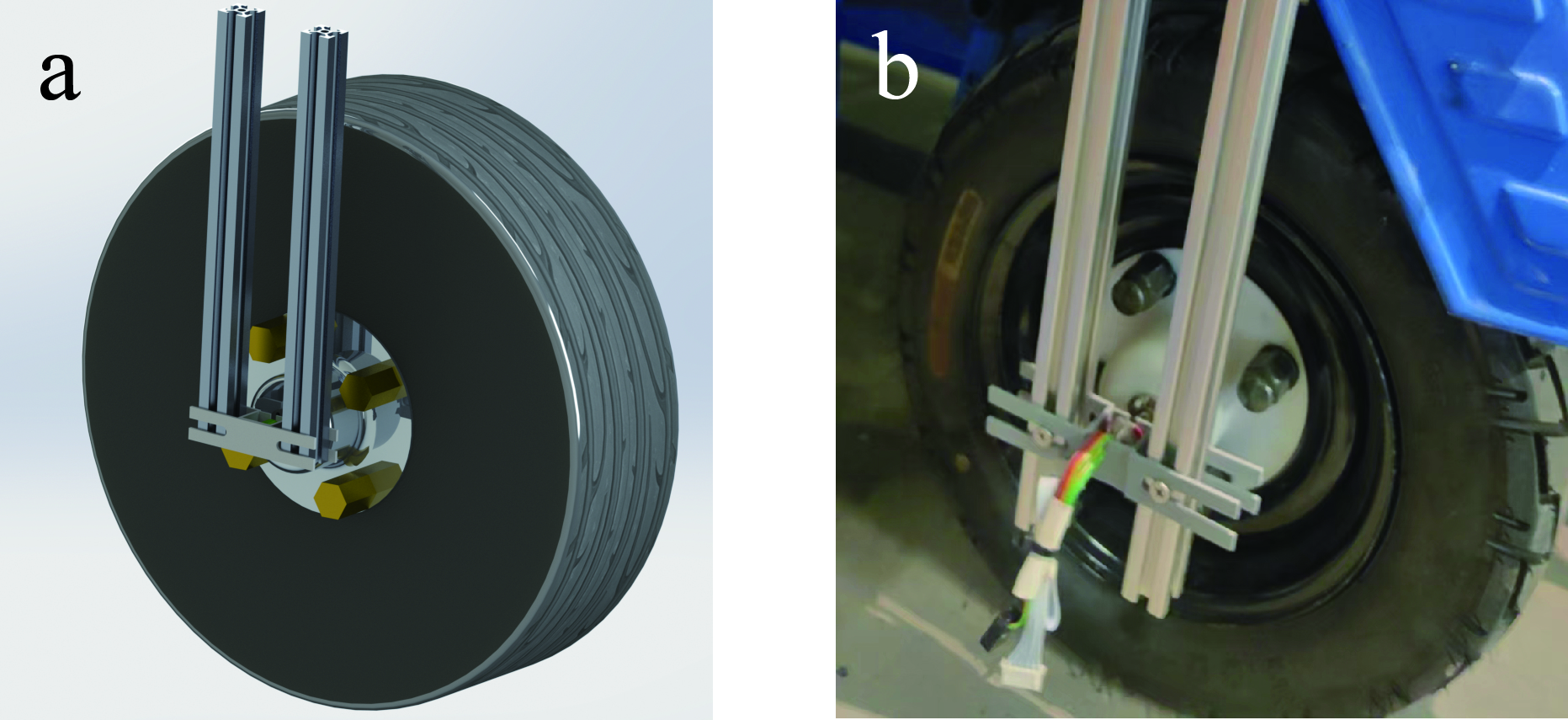}
        \caption{The scheme for encoder fixation and connection. a) Scheme for encoder fixation using aluminum profile and plastic bracket. b) Real picture of encoder and wheel connection.}
    \label{fig:existing_encoder_fixation_plan}
\end{figure}

\begin{figure}[htbp]
     \centering  
     \includegraphics[width=0.5\textwidth]{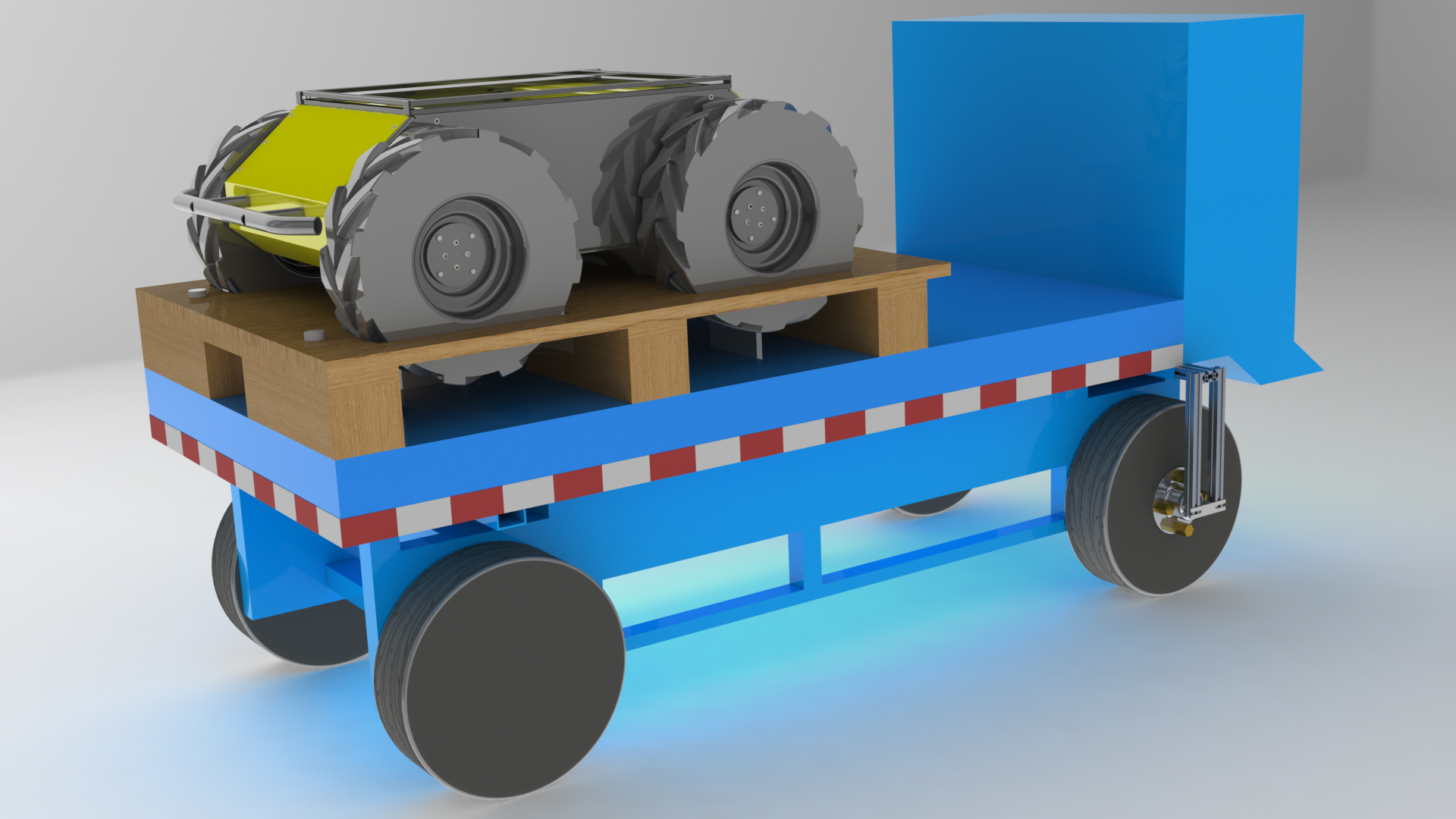}
         \caption{The over all final result of flatbed car modification.}
     \label{fig:final_result}
\end{figure}

The rotation of the wheel is measured relative to the vehicle body reference frame. To ensure accurate acquisition of the wheel rotation values through the encoder's shaft rotation, we need to securely attach the encoder to the vehicle body. We place the encoder in a bracket, which is connected to the flatbed car body using aluminum profiles (Fig.~\ref{fig:existing_encoder_fixation_plan}). This configuration ensures that the rotation of the encoder shaft relative to the encoder body is equal to the rotation of the wheel relative to the vehicle body. The overall final result of the flatbed car modification is shown in Fig.~\ref{fig:final_result}.

\begin{figure}[htbp]
     \centering  
     \includegraphics[width=0.5\textwidth]{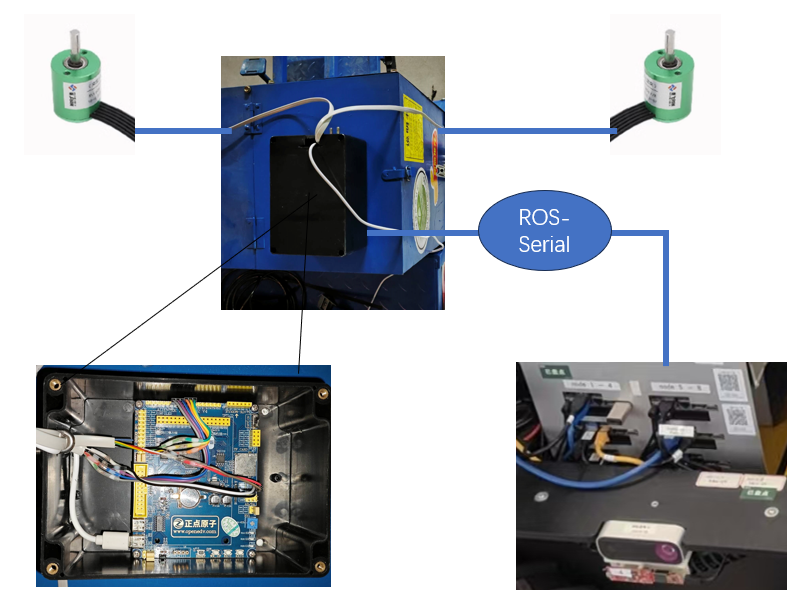}
         \caption{The encoder signals are processed by the MCU and then connected to the cluster via ROS-serial.}
     \label{fig:encoder_mcu}
\end{figure}

The no-name wheel encoders we utilized have 1024 pulses per rotation and are connected to a STM32F103RCT6 development board. We are using the hardware quadrature signal decoding feature of the MCU to get reliable wheel rotation speed and direction readings. The development board is connected via USB to one node of the mapping robot's cluster. There a ROS node is running which is then publishing the encoder readings as ROS messages, such that they may be recorded in ROS bagfiles and/ or used to calculate the robot motion using the odometry model. The connection diagram is shown in Fig.~\ref{fig:encoder_mcu}. All the code is shared in the link provided below.

\subsubsection{Odometry Model}
We model the flatbed car as a two-wheel differential-drive robot, using the non-steered back wheels, that are connected to the drive motor via a differential gear. There are two main parameters: the distance between two wheels, denoted by $L$, the radius of the wheel, denoted by $R$. From the encoder, we get the radian that each wheel travels, multiply it by $R$, get $d_{l,t_k}$ and $d_{r,t_k}$ as the variation of travel of the left and right wheel from time $t_k$ to time $t_{k+1}$. The odometry equation of the platform is as follows,
\begin{align*} x_{t_{k+1}} &= x_{t_k} + \frac{d_{l,t_k} + d_{r,t_k}}{2}  \cos(\theta_{t_k}) \\ 
y_{t_{k+1}} &= y_{t_k} + \frac{d_{l,t_k} + d_{r,t_k}}{2}  \sin(\theta_{t_k}) \\ 
\theta_{t_{k+1}} &= \theta_{t_k} + \frac{d_{r,t_k} - d_{l,t_k}}{L} 
\end{align*}

In order to obtain the coordinates with the origin located on the vehicle, it is necessary to perform the corresponding coordinate system transformation, given that the coordinate system for data collection has its origin positioned on the UGV. When we obtain $l_{0}$ and $h_{0}$ (the difference in distance in the y-direction and z-direction between the rear wheel center and the UGV center, respectively), we can proceed with the coordinate system transformation.The transformation equation from the vehicle coordinate system to the UGV coordinate system is as follows,
\begin{align*} p_{u} &= R_{v}^{u} p_{v} + t_{v}^{u} \\ 
x_{u} &= x_{v} \\ 
y_{u} &= y_{v}-l_{0} \\ 
z_{u} &= z_{v}-h_{0} 
\end{align*}
According to our measurement, $l_{0} = 1.21 m$, $h_{0} = 0.59 m$.

\subsubsection{Odometry Experiment}

\begin{table}[tbp]
\centering
\begin{tabular}{|c|r|r|r|}
\hline 
 & GT & Measured (err)& Optimized (err) \\
\hline
forward & 6.45 m& 6.68 m (3.57\%)& 6.44 m (0.16\%)\\
\hline
forward & 17.37 m& 18.07 m (4.03\%)& 17.43 m (0.35\%)\\
\hline
forward &  10.62 m& 10.93 m (2.92\%)& 10.54 m (0.75\%)\\
\hline
backward & 6.76 m & 6.93 m (2.51\%)& 6.68 m (1.18\%)\\
\hline
circle & 2.63 m & 3.27 m (24.33\%)& 2.61 m (0.76\%)\\
\hline
\end{tabular}
\caption{\textbf{GT} means measuring travel by markers on the ground. \textbf{Measured} and \textbf{Optimized }means calculated odometry with measured and optimized $L, R$ and their errors.}
\label{tab:odom}
\end{table}

To achieve more precise odometry, we implement 5 real-world experiments, including 3 forward, 1 backward, and 1 circle. There we drove the robot a certain distance/ diameter (see Table \ref{tab:odom}) and measured the true distance/ diameter using a tape measure. From Table \ref{tab:odom} we can see that the odometry estimates we get when applying the mesured wheel radius $R$ and wheel baseline $L$ have an err of several percent. 

We optimize our measured $L, R$ with grid search approach. We calculate odometry by different $L, R$ pairs from the product of $L = [0.6, 0.8]$ and $R = [0.15, 0.17]$, where each interval has 50 uniform values. The measured $L = 0.7m$ and $R = 0.1575m$, and the optimized $L = 0.64m$ and $R = 0.164m$, resulting in errors of typically less than one percent.

\begin{figure}[htbp]
    \centering  
        \includegraphics[width=1\linewidth]{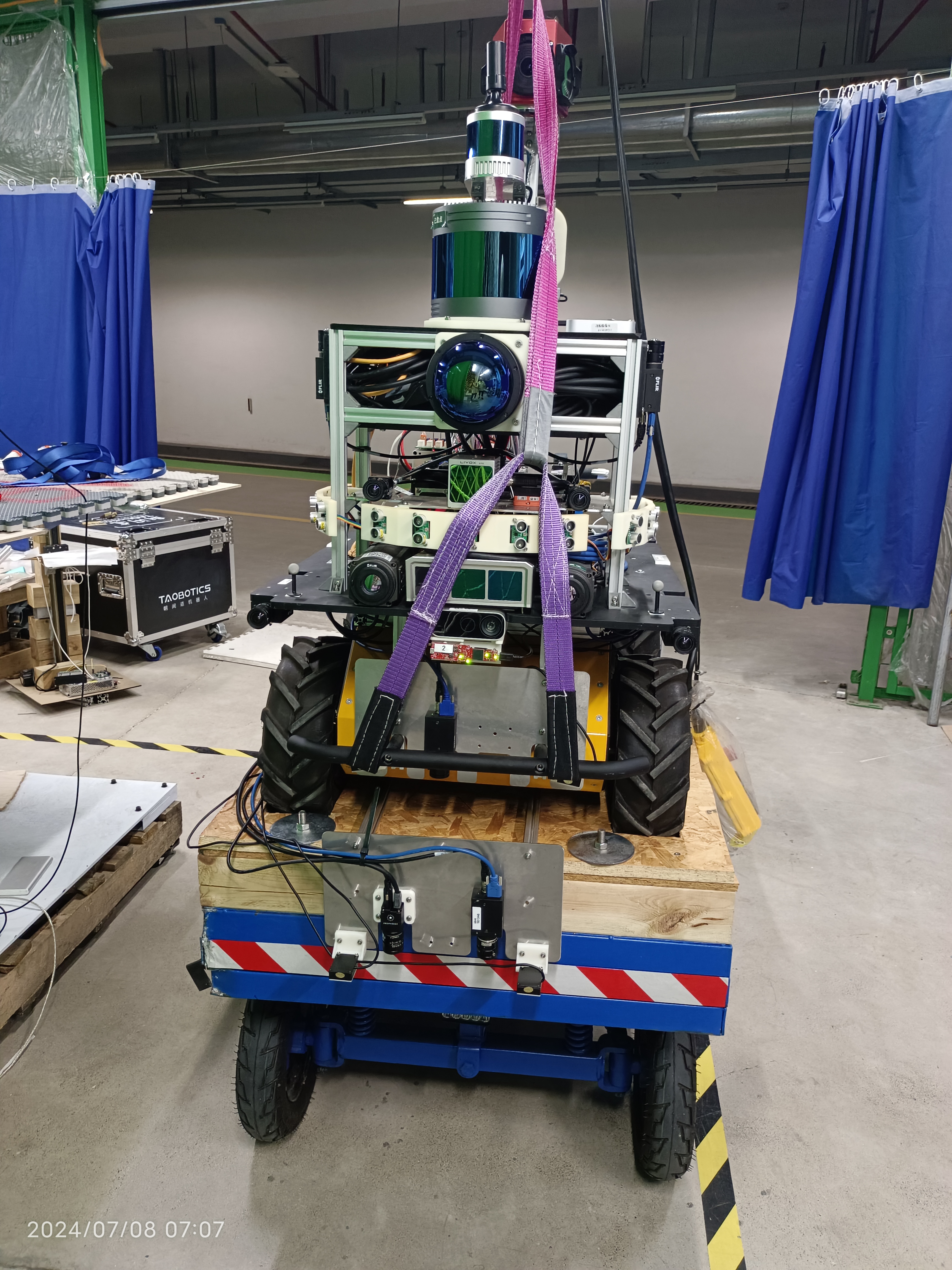}
        \caption{The ShanghaiTech Mapping Robot is lifted with a chain hoist onto the flatbed car for the dataset collection. The external downlooking sensor plate is also visible. }
    \label{fig:rob_on_crane}
\end{figure}

\subsection{Downlooking Sensor Plate}

The ShanghaiTech Mapping Robot features a downlooking sensor plate, in order for algorithms to use this data to calculate even more precise robot motion estimates. But this plate is obstructed by the flatbed car. We have thus designed and installed an external downlooking sensor plate, featuring an event camera, an RGB camera and two non-flickering LED lights, compatible with event cameras. The plate with the devices can be seen in Fig.~\ref{fig:rob_on_crane}. The sensors are connected to the cluster of the mapping robot via USB cables. Additionally, there are hardware synchronization cables running from the IO port of each sensor to the hardware synchronization board of the mapping robot's cluster. Furthermore, there are power lines for the sensors and the LED light connected to the power distribution panel of the robot.


\section{Dataset}
\label{sec:dataset}

With the ShanghaiTech Mapping Robot mounted on the flatbed car we have collected a sample dataset at the beautiful ShanghaiTech University campus. The dataset is 47 minutes and 10.21km long, with an average driving speed of 13km/h. The dataset features, from the downlooking sensor palate, an event camera EVK4 and a downlooking 5MP camera. Next to the collected wheel encoder data, we are sharing the data from the 6-lens Ladybug5+ omnidirectional camera, 6 different LiDARs, the RTK GPS and IMU data. The 884GB dataset can be downloaded here: 
\url{https://robotics.shanghaitech.edu.cn/static/datasets/flatbed/}.\\
This link also provides all the CAD files used in the project, as well as all the code for the collecting and calculating the wheel encoder odometry.

The dataset features the parking garage that is below most of the campus, as well as extensive above-ground traversals of the campus. There are also some nice overlaps with loop-closing opportunities with the ShanghaiTech Outdoor Mapping Robot dataset \cite{xu2024shanghaitech}.

\begin{figure}[htbp]
    \centering  
    \includegraphics[width=1\linewidth]{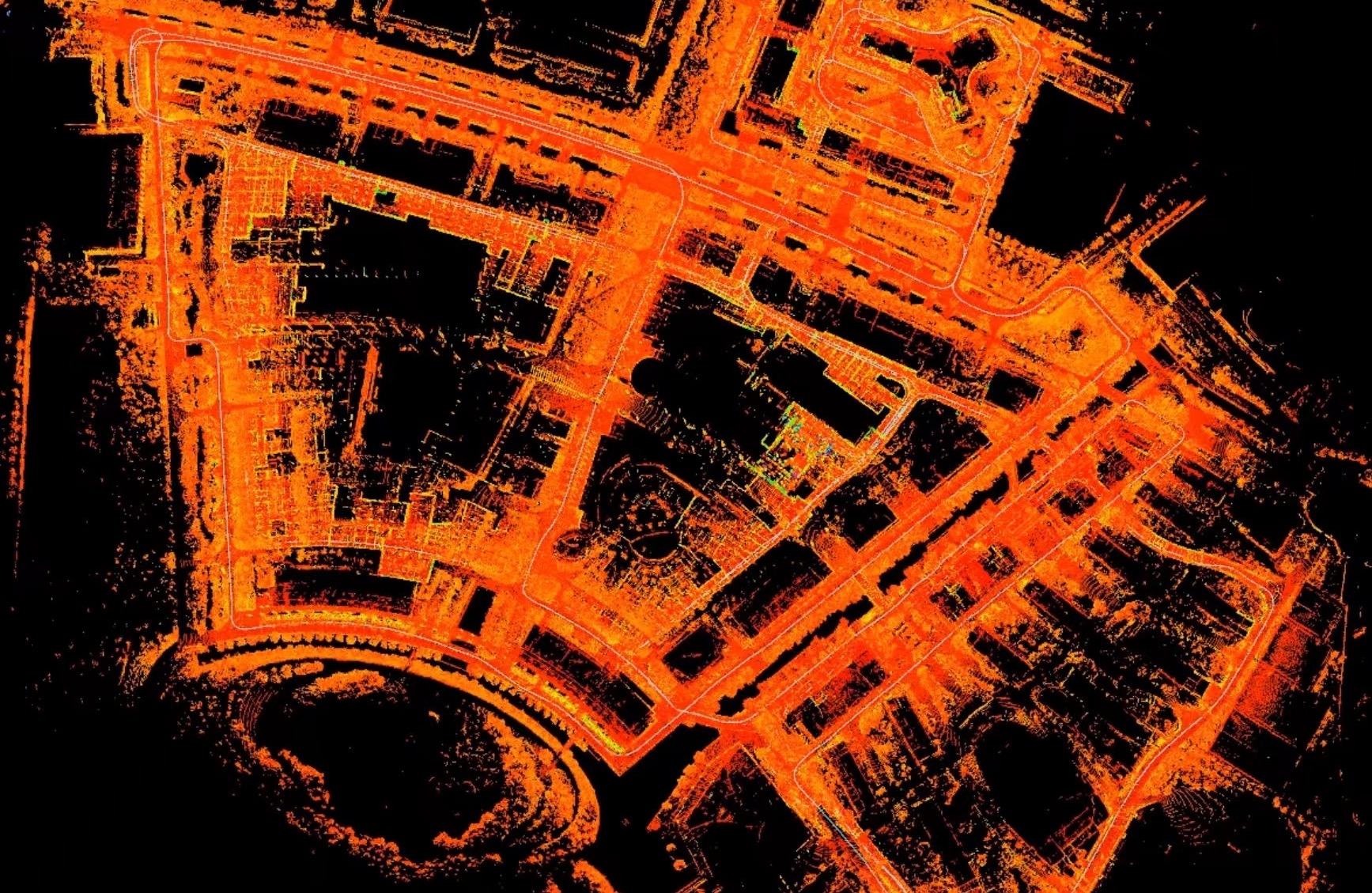}
        \caption{3D point cloud generated with LIO-SAM from the Hesai LiDAR of our dataset.}
    \label{fig:3D_map}
\end{figure}

\subsection{Dataset SLAM Experiment}

From the dataset we employed LIO-SAM \cite{shan2020lio}, a tightly-coupled LiDAR inertial odometry algorithm, to generate a point cloud map from the front robosense 128-beam ruby LiDAR. The point cloud was downsampled to a resolution of 0.1m and has 73 million points - it is depicted in Fig.~\ref{fig:3D_map}.

\section{Conclusions}
\label{sec:conclusions}

The paper presents a high-speed mobile platform for collecting large-scale robotic dataset. The rear-steered flatbed car allows the mapping robot to be placed at the front, maintaining a good field of view for the front-facing sensors and only obstructing a little bit of the back-view. The hydraulic lift even allows to adjust the height of the sensor platform. Furthermore we introduced dual wheel encoders, that can, as our experiments showed, provide reliable platform odometry pose estimates. Our external downlooking sensor plate offers a view of the ground, such that according algorithms may calculate even more precise robot motion estimates. The whole system is working well, as our 10km collected dataset and the point cloud map generated from it demonstrated. In the future we will collect even richer datasets with the system, utilizing all sensors on the ShanghaiTech Mapping Robot, as part of our efforts to collect a very nice ShanghaiTech Robotic Dataset.






\section*{ACKNOWLEDGMENTS}
This work has been partially funded by the Shanghai Frontiers Science Center of Human-centered Artificial Intelligence.
This work was also supported by the Science and Technology Commission of Shanghai Municipality (STCSM), project 22JC1410700 "Evaluation of real-time localization and mapping algorithms for intelligent robots". The experiments of this work were supported by the core facility Platform of Computer Science and Communication, SIST, ShanghaiTech University.




\bibliographystyle{unsrtnat}
\bibliography{ref.bib}

\end{document}